\documentclass[final]{cvpr}

\usepackage{times}
\usepackage{epsfig}
\usepackage{graphicx}
\usepackage{amsmath}
\usepackage{amssymb}
\graphicspath{ {./images/} }
\usepackage{multirow}
\usepackage{float}
\usepackage[noend]{algpseudocode}
\usepackage{algorithmicx}
\usepackage{algorithm}
\usepackage{pifont}
\usepackage{adjustbox}
\usepackage{graphics}
\usepackage{tabularx}
\usepackage{lipsum}
\usepackage{array, boldline, makecell, booktabs}

\usepackage[pagebackref=true,breaklinks=true,colorlinks,bookmarks=false]{hyperref}

\begin{document}

\title{Balance-Oriented Focal Loss with Linear Scheduling\\for Anchor Free Object Detection}

\author{{Hopyong Gil\thanks{Corresponding author, email: thistoopass@mobis.co.kr}\quad Sangwoo Park\quad Yusang Park\quad Wongoo Han\quad Juyean Hong\quad Juneyoung Jung}\\ Hyundai Mobis \\
}

\maketitle

\begin{abstract}
Most existing object detectors suffer from class imbalance problems that hinder balanced performance. In particular, anchor free object detectors have to solve the background imbalance problem due to detection in a per-pixel prediction fashion as well as foreground imbalance problem simultaneously. In this work, we propose Balance-oriented focal loss that can induce balanced learning by considering both background and foreground balance comprehensively. This work aims to address imbalance problem in the situation of using a general unbalanced data of non-extreme distribution not including few shot and the focal loss for anchor free object detector. We use a batch-wise \(\alpha\)-balanced variant of the focal loss to deal with this imbalance problem elaborately. It is a simple and practical solution using only re-weighting for general unbalanced data. It does require neither additional learning cost nor structural change during inference and grouping classes is also unnecessary. Through extensive experiments, we show the performance improvement for each component and analyze the effect of linear scheduling when using re-weighting for the loss. By improving the focal loss in terms of balancing foreground classes, our method achieves AP gains of +1.2 in MS-COCO for the anchor free real-time detector.
\end{abstract}

\section{Introduction}

The object detection is one of the most actively researched task in computer vision and has been widely used in various applications. Object detectors based on deep learning are mainly divided into single stage {\cite{redmon2017yolo9000,liu2016ssd,law2018cornernet,zhou2019objects,tian2019fcos} and two stage {\cite{ren2015faster,he2017mask,cai2018cascade}} depending on whether there is the additional refinement stage for proposal boxes. Although single stage object detectors have inferior performance compared with two stage object detectors in general, they have been extensively used in many applications as on-device algorithm because of advantages in speed and memory. For single stage detectors, there are two types of detector, anchor-based {\cite{redmon2017yolo9000,liu2016ssd}, and anchor-free {\cite{law2018cornernet,zhou2019objects,tian2019fcos,duan2019centernet}}. Recently, various ways for anchor free detector have been proposed. This is because the anchor free detector can reduce complex operations and hyper-parameters search in anchor-based detectors, and has a relatively simple and intuitive structure. So it has advantages on edge devices.

\begin{figure}
    \begin{center}
    \includegraphics[width=0.48\textwidth]{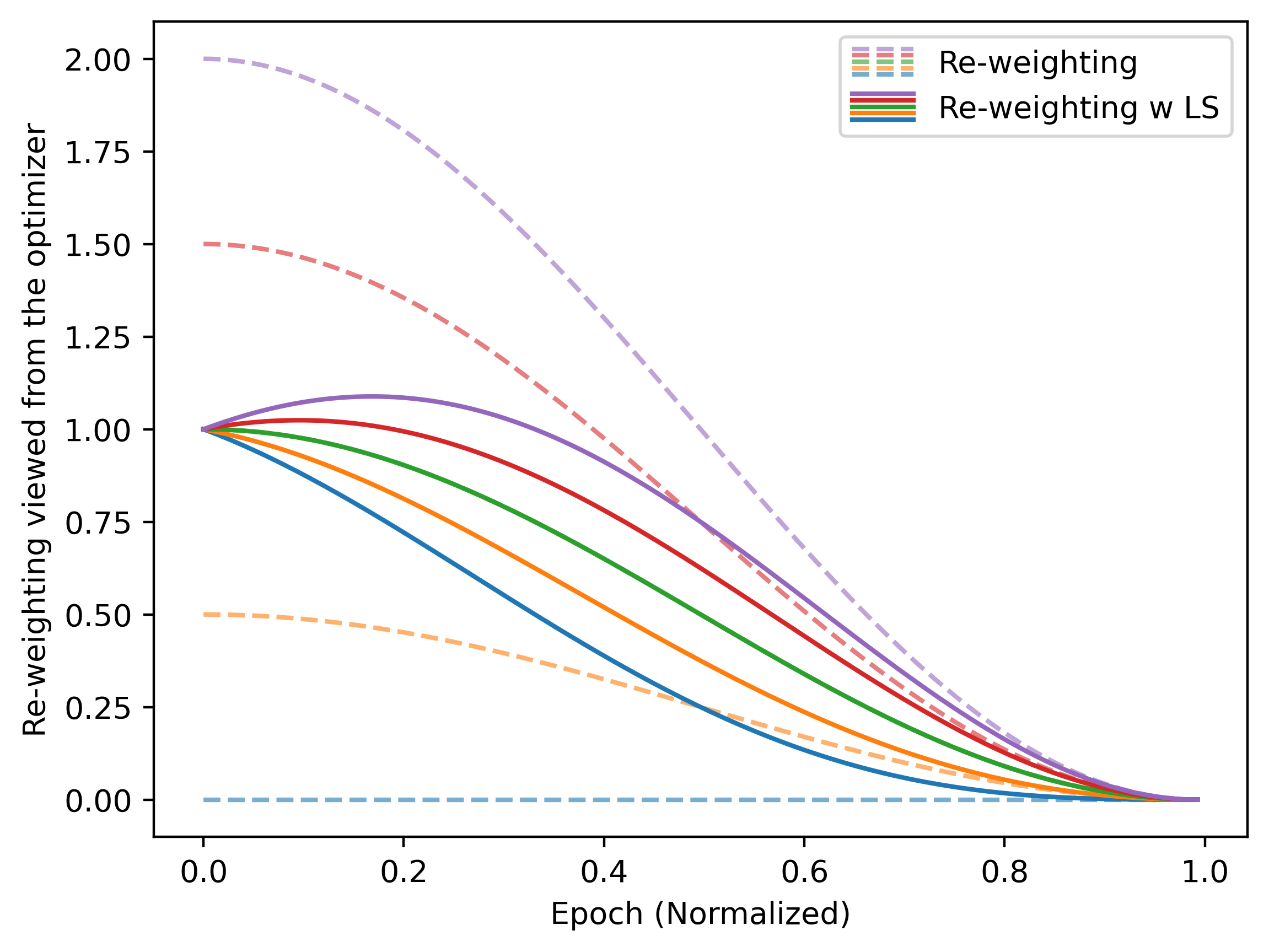}
    \caption{Visualization for the effect of re-weighting without (dashed line) and with (solid line) linear scheduling (LS) from the perspective of the optimizer. It is shown by multiplying a learning rate by re-weighting factors that purple, red, green, yellow and blue stand for 2.0, 1.5, 1.0, 0.5 and 0.0, respectively. For simplicity, cosine annealing with initial value = 1 is used as a learning rate scheduler.}
    \label{fig:fig1}
    \end{center}
\end{figure}

These object detectors basically take a supervised learning using annotation data. The imbalance problem between foreground classes is an inevitable {\cite{oksuz2020imbalance}} when data is acquired. Recently, many methods have been proposed to deal with this imbalance problem between foreground classes {\cite{cui2019class,tan2020equalization,li2020overcoming,wang2020devil,cao2019learning}}. They only performed experiments with the two stage detectors {\cite{he2017mask,cai2018cascade}} to increase an absolute performance. These methods aim to solve the class imbalance problem on the classifier of the second stage that receives proposals with the background filtered from the first stage as inputs. Therefore, they can focus on the class imbalance problem between foregrounds. However, many anchor free detectors {\cite{law2018cornernet,zhou2019objects,tian2019fcos,duan2019centernet} structurally encounter severe foreground-background class imbalance during training because they detect objects directly on a feature map in a per-pixel prediction fashion. Thus, it may not work well when using the existing methods {\cite{tan2020equalization,li2020overcoming}} that tried to solve only the foreground class imbalance problem in such a situation where the background class imbalance is not so severe.

Also, most of them {\cite{tan2020equalization,li2020overcoming,wang2020devil}} were experimented on data {\cite{gupta2019lvis,kuznetsova2018open}} extremely unbalanced including few shots images. But many object detectors in applications rarely use such an extremely unbalanced data. Except for special cases, they use unbalanced data of general level {\cite{lin2014microsoft,yu2020bdd100k,neuhold2017mapillary}} because common applications (\eg, safety-critical system) require stable performance for all classes. In practical, even if it is rare class, it is not so difficult to obtain the minimum quantity of data (\eg, more than 100 images) {\cite{lin2014microsoft}}. It is much more difficult and popular issue to get data that is perfectly balanced. Existing methods on unbalanced data require grouping between classes {\cite{li2020overcoming}} or focus on the suppression effect of tail classes {\cite{tan2020equalization}} because they developed mainly for data with extreme class imbalance {\cite{gupta2019lvis}}. The method that is directly applicable to the class imbalance problem on a general unbalanced data is understudied in the literature.

In this situation, Focal loss {\cite{lin2017focal}} has became one of the most popular solution for the class imbalance problem by down-weighting the loss of well-classified examples in order to solve the imbalance caused by the easy negatives overflowing in the single stage detector. However, it has a limitation that it does not handle class imbalance problem between foregrounds efficiently. Class balanced loss {\cite{cui2019class}} proposed a theoretical framework by approaching class imbalance problem through the effective number of samples, and showed that it is possible to apply their method to focal loss as it can be viewed as a class balanced version of the \(\alpha\)-balanced term in focal loss. But they do not optimize the \(\alpha\) term of the balanced focal loss for each batch and for each epoch finely.

In this paper, we propose a Balance-oriented focal loss to take into account both background and foreground imbalance. By considering the distribution of the class instances in the dataset level and batch level, we improve \(\alpha\) term of the focal loss {\cite{lin2017focal}} for each image in a well-balanced manner. Furthermore, this re-weighting is not applied uniformly during entire training step, but is linearly changed as illustrated in Figure {\ref{fig:fig1} (solid line). It can make optimization process stable in a fine-tuning manner by not moving parameters of the original model very far at the beginning of training. Our method is a simple and practical solution to general unbalanced data. It does not require dividing groups between classes {\cite{tan2020equalization,li2020overcoming}} or additional learning cost {\cite{wang2020devil}} or post-processing when inference. Finally, we show Balance-oriented focal loss can improve the performance of anchor free real-time detector and analyze the effectiveness of each component through extensive experiments.

\section{Related Works}

{\bf Anchor Free Detectors.} The anchor free detectors {\cite{law2018cornernet,zhou2019objects,tian2019fcos,tian2019fcos,duan2019centernet} have been developed actively in recent years. They have a relatively simple and intuitive structure by reducing the complicated operation required in an anchor-based object detector.
CornerNet {\cite{law2018cornernet} showed that it is possible to obtain high performance with point-based detection, and played a pioneering role as an anchor free detector. CenterNet {\cite{zhou2019objects}} was developed based on CornerNet by replacing the corner points with the center point of the box. It can detect objects efficiently by eliminating grouping corners and extra operations from CornerNet. FCOS {\cite{tian2019fcos}} is also anchor free detector in a per-pixel prediction fashion. It predicts the distance to each side of the box from all pixels on the feature and takes post-processing with centerness estimation to reduce false positives. For simplicity, we have mainly experimented our method on CenterNet as a representative anchor free detector using focal loss {\cite{lin2017focal}}.

{\bf Learning Imbalanced Data.} Data is notably important in object detection task. In practice, almost all data has a class imbalance problem. For this reason, many methods have been proposed to solve the class imbalance problem between foreground classes. Balanced group softmax {\cite{li2020overcoming}} improved the performance on long tailed dataset by using group-wise training for the classifier, which divides classes into several groups by frequency. Equalization loss {\cite{tan2020equalization}} overcame performance degradation caused by discouraging gradients of tail categories on long tailed data. However, these methods basically assume extreme class imbalance conditions and should divide classes into several groups. In addition, it is unclear to divide several groups between classes when there is no extreme class imbalance on the data. SimCal {\cite{wang2020devil}} introduced calibration training using balanced bi-level sampling, but it has a disadvantage as it requires additional training cost for fine-tuning and extra head for dual inference. Therefore, we aim at general unbalanced data with non-extreme class imbalance and try to solve this problem by improving the loss term without additional computation cost or changing the structure of the model.

{\bf Re-Weighting Methods.}
Training strategies on class unbalanced data with re-weighting methods have been presented in a variety of ways. Focal loss {\cite{lin2017focal}} greatly improved the performance by down-weighing the loss of well-classified samples to solve the imbalance caused by the overflow of easy negatives in dense detector. However, it can not handle class imbalance problem between foregrounds aptly. Class balanced loss {\cite{cui2019class}} proposed a theoretical framework by approaching class imbalance problem through an effective number of samples, and showed that it can be applied to focal loss as class balanced version of \(\alpha\)-balanced term. This class balanced focal loss is the motivation of our method. And we have adjusted the \(\alpha\) term of the balanced focal loss for each batch and each epoch flexibly.

 \begin{table*}
     \begin{center}
     \begin{tabularx}{1.0\linewidth}{l|*{3}{>{\centering\arraybackslash}X|}*{3}{>{\centering\arraybackslash}X}|*{3}{>{\centering\arraybackslash}X}} & WF & BB & LS & $AP$ & $AP_{50}$ & $AP_{75}$ & $AP_S$ & $AP_M$ & $AP_L$\\
     \Xhline{1pt} 
     \multirow{1}{*}{Baseline} & \multirow{1}{*}{} & \multirow{1}{*}{} & \multirow{1}{*}{} & \multicolumn{1}{c}{26.4}&\multicolumn{1}{c}{43.4}&\multicolumn{1}{c|}{27.4} & \multicolumn{1}{c}{9.2}&\multicolumn{1}{c}{27.2}&\multicolumn{1}{c}{41.5}\\ 
     \hline
     \multirow{1}{*}{+WF} & \multirow{1}{*}{\checkmark} & \multirow{1}{*}{} & \multirow{1}{*}{} & \multicolumn{1}{c}{25.9}&\multicolumn{1}{c}{43.8}&\multicolumn{1}{c|}{26.2} & \multicolumn{1}{c}{10.3}&\multicolumn{1}{c}{26.1}&\multicolumn{1}{c}{40.8}\\
     \hline 
     \multirow{1}{*}{+WF +LS} & \multirow{1}{*}{\checkmark} & \multirow{1}{*}{} & \multirow{1}{*}{\checkmark} & \multicolumn{1}{c}{27.2}&\multicolumn{1}{c}{45.1}&\multicolumn{1}{c|}{28.1} & \multicolumn{1}{c}{10.2}&\multicolumn{1}{c}{28.1}&\multicolumn{1}{c}{\bf42.4}\\
     \hline 
     \multirow{1}{*}{+BB} & \multirow{1}{*}{} & \multirow{1}{*}{\checkmark} & \multirow{1}{*}{} & \multicolumn{1}{c}{26.3} & \multicolumn{1}{c}{42.8} & \multicolumn{1}{c|}{27.4} & \multicolumn{1}{c}{9.5} & \multicolumn{1}{c}{26.8} & \multicolumn{1}{c}{41.3}\\
     \hline
     \multirow{1}{*}{+BB +LS} & \multirow{1}{*}{} & \multirow{1}{*}{\checkmark} & \multirow{1}{*}{\checkmark} & \multicolumn{1}{c}{27.1} & \multicolumn{1}{c}{44.8} & \multicolumn{1}{c|}{28.3} & \multicolumn{1}{c}{\bf10.5} & \multicolumn{1}{c}{28.0} & \multicolumn{1}{c}{42.1}\\
     \hline
     \multirow{1}{*}{+WF +BB +LS (Ours)} & \multirow{1}{*}{\checkmark} & \multirow{1}{*}{\checkmark} & \multirow{1}{*}{\checkmark} & \multicolumn{1}{c}{\bf27.6} & \multicolumn{1}{c}{\bf45.6} & \multicolumn{1}{c|}{\bf28.7} & \multicolumn{1}{c}{10.3} & \multicolumn{1}{c}{\bf28.3} & \multicolumn{1}{c}{42.1}
     \end{tabularx}
     \end{center}
     \caption{Results on different components. All models are evaluated on val2017 in MS-COCO. The evaluation metric is box AP across IoU threshold from 0.5 to 0.95 over all categories as following COCO evaluation process. 'WF', 'BB' and 'LS' stand for weighting factor, batch-wise balancing and linear scheduling, respectively.}
     \label{tab:hptab1}
 \end{table*}
\section{Preliminary}
{\bf Class Balanced Focal Loss.} Class balanced focal loss {\cite{cui2019class}} theoretically approached re-weighting strategy through the effective number of samples to solve the class imbalance problem. It can be written as: 
 \begin{equation}
     \label{eq:hp1}
     \mathrm{\textbf{CB}}_{\mathrm{focal}}(y) = -\alpha_y{\sum_{i=1}^{C} (1 - p_i^{t})^{\gamma}log(p_i^{t})}
 \end{equation}
 where
   \begin{equation}
   \label{eq:hp2}
 {\alpha_y = \frac{1-{\beta}}{1-{\beta}^{n_y}}},
   \end{equation}
  \begin{equation}
     \label{eq:hp3}
        p_i^{t}=
        \begin{cases}
        p_i     & \text{if }\ y=1\\
        1-p_i   & \text{otherwise}
        \end{cases}
  \end{equation} where label $y \in \{1, 2, . . . , C\}$ and $C$ is the total number of classes. \(p_i\) is the probability of detection result for class \(i\) and \(n_y\) is the number of samples in the ground-truth class \(y\). \(\alpha_y\) is weighting factor for class balanced term.
   \begin{equation}
     \label{eq:hp4}
     N = \lim_{n\to\infty} {\sum_{j=1}^n \beta^{j-1}} = 1/(1 - \beta).
 \end{equation}
  \(\beta\) is hyper-parameter related to the expected total volume \(N\) where \(n\) is the number of instances as Equation \ref{eq:hp4}. They showed that the class balanced focal loss works well on the long-tailed classification dataset.

{\bf Approximation of Weighting Factor.} We take the approximation for \(\alpha_y\) to exclude the dense search for hyper-parameter \(\beta\), since our goal is finding how to apply it stably rather than finding the optimal weighting factor itself. In general, the number of objects is larger than the number of images by one order of magnitude in object detection task. So we assume that N is very large. Then if N is very large, according to Equation \ref{eq:hp4}, the \(\beta\) approaches 1, which means that the weighting factor approaches the inverse class frequency. Hence, the weighting factor is approximated by increasing or decreasing from \(1/C\) that is equally distributed case among all classes according to the ratio of the number of instances of the class as: 
 \begin{equation}
     \label{eq:hp5}
     {\alpha_y = \frac{n}{C \times n_y}}.
 \end{equation}

\section{Balance-Oriented Focal Loss with Linear Scheduling}
\subsection{Re-Weighting on Focal Loss}
Our goal is to find the re-weighting method suitable for an anchor free detector using focal loss. And we expect that the method of applying \(\alpha\) for each sample as Equation \ref{eq:hp1} is ineffective to produce sufficient balancing effect for anchor free detectors that the number of samples of background class is overwhelmingly larger than that of the foreground classes. Therefore we use the same \(\alpha\) for all samples, but apply it differently for each class in focal loss as:
  \begin{equation}
     \label{eq:hp9}
     \mathrm{\Grave{\textbf{CB}}_{\mathrm{focal}}}(y) = -{\sum_{i=1}^{C} \alpha_i(1 - p_i^{t})^{\gamma}log(p_i^{t})}.
 \end{equation}
 
\subsection{Linear Scheduling}
In the previous section, we have determined weighting factor. But applying re-weighting method directly can degrade the performance of the original detector because it forcibly changes the geometry of the loss surface. Therefore, we take a linear scheduling technique that gradually increases the intensity of weighting factor as training progresses by the following: 
 \begin{equation}
     \label{eq:hp6}
     {\hat{\alpha_i} = 1+\lambda}(\alpha_i - 1)
 \end{equation}
 where $\lambda \in [0, 1]$ is the normalized epoch and \(\hat{\alpha_i}\) is weighting factor with linear scheduling. The network is trained as usual by setting weighting factor to 1 at the start of training. And then, the loss for each class is gradually scaled in proportion to the inverse class frequency as it is being trained and the learning rate decreases. We visualize the effect of linear scheduling in the optimizer in Figure \ref{fig:fig1} by weighting class balanced term to cosine annealing learning rate schedule. It shows that our method can apply the re-weighting method in a  fine-tuning manner while balancing foreground classes during training. Therefore, our method has an advantage in that it optimizes the loss in a class balanced manner without degrading the original performance.

\subsection{Batch-Wise Balancing}
Re-weighting has applied to solve the dataset level class imbalance, but the batch level class imbalance has not been considered yet. Thus, we also take into account the instance imbalance at the batch level where the parameters of the model are updated. The class balanced term considering the batch level class imbalance is as:
 \begin{equation}
     \label{eq:hp7}
     {\hat{w}_{i,k} = \hat{\alpha_i} \hat{\eta}^{n_{i,k}}}
 \end{equation}
 where \(n_{i,k}\) is the number of samples of the ground-truth class \(i\) in the single batch \(k\) which is the index of an image. And \(\eta\) is the hyper-parameter with a range of [0, 1] because it aims at the effect of down-weighting the loss as the number of objects increases. \(\hat{hat}\) means to apply linear scheduling technique as Equation \ref{eq:hp6}. We show that the effectiveness of batch-wise balancing in Section \ref{sec:ablation}. It can improve the performance by itself without dataset level class balancing.
 Finally, we define Balance-Oriented Focal Loss with Linear Scheduling (BOFL) as:
  \begin{equation}
     \label{eq:hp8}
     \mathrm{\textbf{BO}}_{\mathrm{focal}}(y, k) = -{\sum_{i=1}^{C} \hat{w}_{i,k} (1 - p_i^{t})^{\gamma}log(p_i^{t})}.
 \end{equation}

 \begin{table}[!t]
     \begin{center}
     \begin{tabular}{c|ccc}
     $\eta$ & $AP$ & $AP_{50}$ & $AP_{75}$\\
     \Xhline{1pt}
 \multirow{1}{*}{\centering0.6}          & \multicolumn{1}{c}{27.5}&\multicolumn{1}{c}{45.3}&\multicolumn{1}{c}{28.6}\\ 
 \hline
 \multirow{1}{*}{\centering0.7}         & \multicolumn{1}{c}{27.4}&\multicolumn{1}{c}{45.1}&\multicolumn{1}{c}{28.4}\\
 \hline                           
 \multirow{1}{*}{\centering0.8}          & \multicolumn{1}{c}{\bf27.6}&\multicolumn{1}{c}{\bf45.6}&\multicolumn{1}{c}{\bf28.7}\\
 \hline                         
 \multirow{1}{*}{\centering0.9}         & \multicolumn{1}{c}{27.1}&\multicolumn{1}{c}{45.1}&\multicolumn{1}{c}{28.1}\\
     \end{tabular}
     \end{center}
     \caption{Results for varying \(\eta\).}
     \label{tab:hptab2}
 \end{table}

 
\section{Experiments}
The proposed method is evaluated on MS-COCO dataset {\cite{lin2014microsoft}}, which is one of the most general unbalanced dataset in the object detection task. MS-COCO dataset has 80 classes and contains 118k training images (train2017), 5k validation images (val2017) and 20k hold-out testing images (test-dev). All models are trained on train2017 and tested on val2017 unless noted.

 \begin{table}[!t]
     \begin{center}
     \begin{tabular}{l|ccc}
     Backbone & $AP$ & $AP_{50}$ & $AP_{75}$\\
     \Xhline{1pt}
      \multirow{1}{*}{Mobilenet v3}          & \multicolumn{1}{c}{26.2}&\multicolumn{1}{c}{44.2}&\multicolumn{1}{c}{27.0}\\ 
 \multirow{1}{*}{+BOFL}          & \multicolumn{1}{c}{\bf27.3}&\multicolumn{1}{c}{\bf45.8}&\multicolumn{1}{c}{\bf28.2}\\ 
 \hline
  \multirow{1}{*}{Resnet 18}         & \multicolumn{1}{c}{30.0}&\multicolumn{1}{c}{47.4}&\multicolumn{1}{c}{31.5}\\
 \multirow{1}{*}{+BOFL}         & \multicolumn{1}{c}{\bf30.4}&\multicolumn{1}{c}{\bf48.1}&\multicolumn{1}{c}{\bf32.1}\\
 \hline  
  \multirow{1}{*}{Resnet 101}          & \multicolumn{1}{c}{35.0}&\multicolumn{1}{c}{53.5}&\multicolumn{1}{c}{37.1}\\
 \multirow{1}{*}{+BOFL}          & \multicolumn{1}{c}{35.0}&\multicolumn{1}{c}{\bf53.7}&\multicolumn{1}{c}{\bf37.3}\\
 \hline  
   \multirow{1}{*}{DLA 34}          & \multicolumn{1}{c}{37.2}&\multicolumn{1}{c}{55.0}&\multicolumn{1}{c}{40.2}\\
 \multirow{1}{*}{+BOFL}          & \multicolumn{1}{c}{\bf37.6}&\multicolumn{1}{c}{\bf55.8}&\multicolumn{1}{c}{\bf40.6}\\
     \end{tabular}
     \end{center}
     \caption{Results on different backbones.}
     \label{tab:hptab5}
 \end{table}

\subsection{Implementation Details}
We use CenterNet {\cite{zhou2019objects}} to conduct experiments. It predicts the center of an object for every pixel on the feature map, and then estimates the width and height at the center. It is implemented based on CornerNet {\cite{law2018cornernet}, but is faster by removing complex post-processing. We train with a batch size of 128 on NVIDIA V100 4 GPU for 140 epochs. The initial learning rate is 0.0005, and the initial 500 iterations are set as warmup {\cite{goyal2017accurate}}. We take cosine annealing {\cite{loshchilov2016sgdr}} as learning rate schedule with T = total epochs, and the learning rate decreased until 0. The input size is 512 x 512 and the weight decay is 0.0001. BatchNorm layer {\cite{ioffe2015batch}} is added after the activation layer in the head, and the deformable convolution {\cite{dai2017deformable}} in the neck is replaced with normal convolution operation when the Mobilenet {\cite{sandler2018mobilenetv2,howard2019searching}} series, which is known to be optimized for the embedded environment, is used as a backbone. Augmentation and other settings are used in the same way as the original {\cite{zhou2019objects}} . Unless otherwise specified, our experiments are conducted with Mobilenet v2 {\cite{sandler2018mobilenetv2}}.

\subsection{Ablation Studies} \label{sec:ablation} 

{\bf Weighting Factor.}
We report the effects of different components through experiments. When only the weighting factor at the dataset level is added as a class balanced term to baseline, AP decreases as shown in Table \ref{tab:hptab1}. This is the case of Figure \ref{fig:fig1} (dashed line). The reason for this is that the class weighting with extreme values (\eg, 1\(\gg\) or 1\(\ll\)) may cause a bad initialization and it makes optimization process unstable. After adding linear scheduling, we get AP gains of +0.8 to baseline.

{\bf Batch-Wise Balancing.}
When class balancing at the batch level is added to baseline, AP decreases slightly as shown in Table \ref{tab:hptab1}. By adding linear scheduling, the performance of +0.7 AP is improved. This shows that it is possible to improve performance with only class balanced terms at the batch level.

{\bf The Effect of \(\eta\).}
In our method, there is a hyper-parameter \(\eta\) in the class balanced term for the batch level balancing. According to Table \ref{tab:hptab2}, \(\eta\) = 0.8 gives the best performance. Since this is affected by the number of instances per image, the optimal value will depend on the dataset. Unless otherwise noted, we set \(\eta\) to 0.8 for all experiments.

{\bf Backbone.}
  We also experiment effectiveness of our method on different backbones as shown in Table \ref{tab:hptab5}. For Mobilenet v3 {\cite{howard2019searching}}, the performance of +1.1 AP is improved. For Resnet {\cite{he2016deep}} and DLA {\cite{yu2018deep}}, we run inference while maintaining the resolution of the image as following the original setting. We only change the learning rate schedule from step decay to cosine annealing {\cite{loshchilov2016sgdr}}. For Resnet 18 and DLA 34, they achieve AP gains of +0.4. For Resnet 101, there is no improvement. The result shows that the performance improvement of our proposed method is remarkable on the embedded-friendly model {\cite{sandler2018mobilenetv2,howard2019searching}}.

 \begin{table}[!t]
     \begin{center}
     \begin{tabular}{l|ccc}
      \centering DRW& $AP$ & $AP_{50}$ & $AP_{75}$\\
     \Xhline{1pt}
  \multirow{1}{*}{1}         & \multicolumn{1}{c}{\bf27.6}&\multicolumn{1}{c}{\bf45.7}&\multicolumn{1}{c}{\bf28.7}\\
 \hline   
 \multirow{1}{*}{5}         & \multicolumn{1}{c}{27.6}&\multicolumn{1}{c}{45.6}&\multicolumn{1}{c}{28.7}\\
 \hline                           
 \multirow{1}{*}{15}          & \multicolumn{1}{c}{27.3}&\multicolumn{1}{c}{45.0}&\multicolumn{1}{c}{28.1}\\
 \hline                         
 \multirow{1}{*}{25}         & \multicolumn{1}{c}{27.3}&\multicolumn{1}{c}{45.1}&\multicolumn{1}{c}{28.3}\\
     \end{tabular}
     \end{center}
     \caption{Results for varying deferred epochs in DRW.}
     \label{tab:hptab3}
 \end{table}

 \begin{table}[!t]
     \begin{center}
     \begin{tabular}{l|ccc}
      \centering & $AP$ & $AP_{50}$ & $AP_{75}$\\
     \Xhline{1pt}
 \multirow{1}{*}{constant}          & \multicolumn{1}{c}{25.9}&\multicolumn{1}{c}{43.8}&\multicolumn{1}{c}{26.2}\\ 
 \hline
 \multirow{1}{*}{deferred}         & \multicolumn{1}{c}{26.0}&\multicolumn{1}{c}{43.2}&\multicolumn{1}{c}{26.6}\\
 \hline                           
 \multirow{1}{*}{linear}          & \multicolumn{1}{c}{27.4}&\multicolumn{1}{c}{45.4}&\multicolumn{1}{c}{28.1}\\
  \hline                           
 \multirow{1}{*}{ours}          & \multicolumn{1}{c}{\bf27.6}&\multicolumn{1}{c}{\bf45.6}&\multicolumn{1}{c}{\bf28.7}\\
     \end{tabular}
     \end{center}
     \caption{Results on different schedules for re-weighting.}
     \label{tab:hptab4}
 \end{table}

 \begin{table}[!t]
     \begin{center}
     \begin{tabular}{l|ccc}
      \centering & $AP$ & $AP_{50}$ & $AP_{75}$\\
     \Xhline{1pt}
 \multirow{1}{*}{step decay}          & \multicolumn{1}{c}{26.4}&\multicolumn{1}{c}{43.6}&\multicolumn{1}{c}{27.1}\\ 
  \multirow{1}{*}{+BOFL}          & \multicolumn{1}{c}{\bf26.9}&\multicolumn{1}{c}{\bf44.5}&\multicolumn{1}{c}{\bf27.8}\\ 
 \hline
 \multirow{1}{*}{cosine}         & \multicolumn{1}{c}{26.4}&\multicolumn{1}{c}{43.4}&\multicolumn{1}{c}{27.4}\\
  \multirow{1}{*}{+BOFL}         & \multicolumn{1}{c}{\bf27.6}&\multicolumn{1}{c}{\bf45.6}&\multicolumn{1}{c}{\bf28.7}\\
     \end{tabular}
     \end{center}
     \caption{Results on different learning rate scheduler.}
     \label{tab:hptab6}
 \end{table}
 
\subsection{The Effect of DRW}
We analyze the effect of the scheduling type for class balanced term. In {\cite{cao2019learning}}, they proposed deferred re-weighting (DRW) training schedule which helps improve the performance by deferring the start of re-balancing for several iteration. It can prevent parameters of the model from being too largely distracted due to re-weighting in the beginning of training. We take DRW because it has a similar philosophy to our method in a fine-tuning manner for class balancing. As shown in Table \ref{tab:hptab3}, DRW is helpful in improving the performance. We use 5 epochs, which lead almost the best performance, as the default value of deferred epochs for all experiment unless noted.
We also experiment with the case of using a constant without the deferred schedule, using a constant after deferred epochs, using linear scheduling without the deferred schedule and using linear scheduling after deferred epochs for the weighting factor. We visualize them in Figure \ref{fig:fig2} as solid, dotted, dashed and dash-dotted line, respectively. As shown in Table \ref{tab:hptab4}, linear scheduling after deferred epochs can produce the best performance.

\begin{figure}
    \begin{center}
    \includegraphics[width=0.47\textwidth]{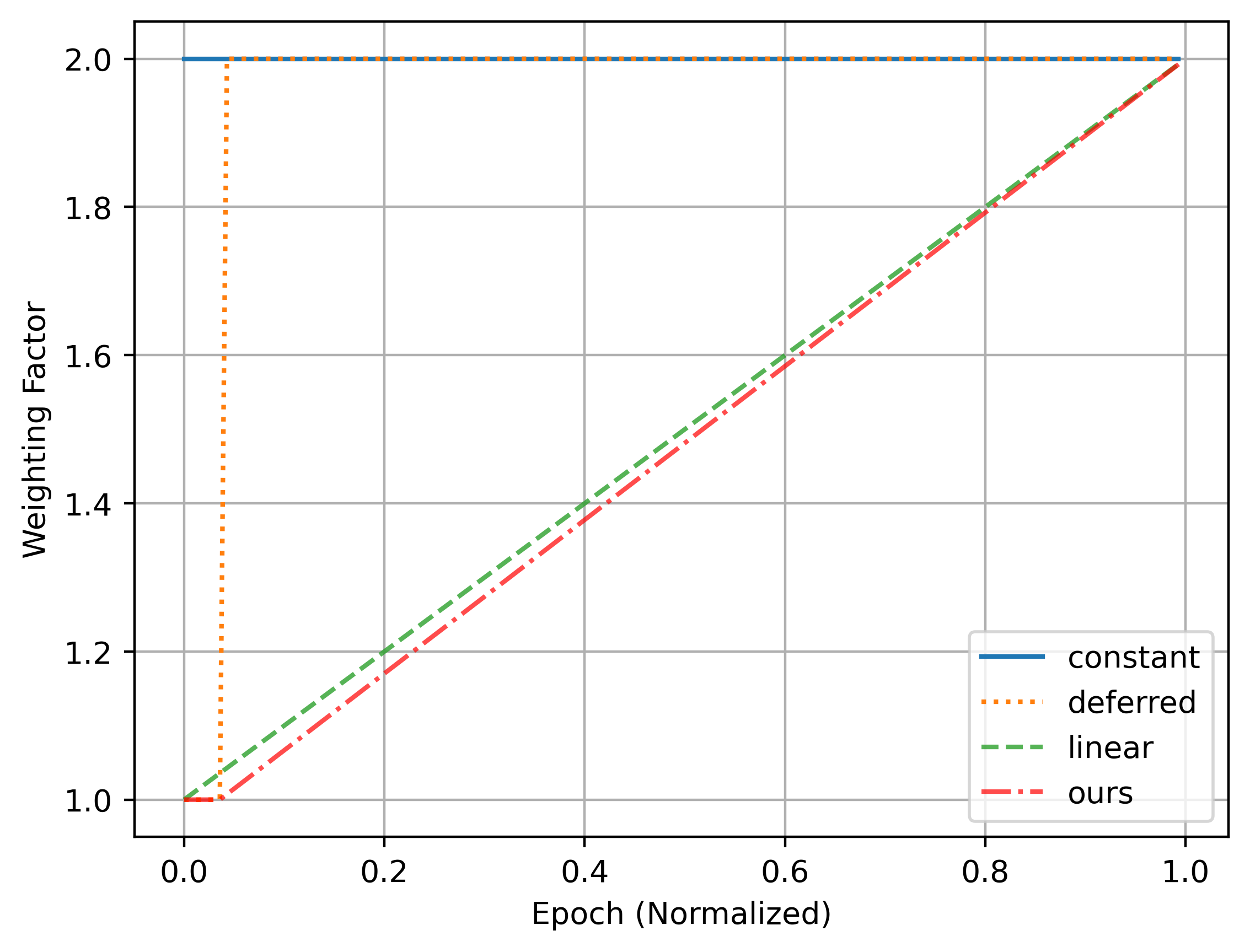}
    \caption{The different schedules for re-weighting when weighting factor is 2.}
    \label{fig:fig2}
    \end{center}
\end{figure}

 \begin{table}[!t]
     \begin{center}
     \begin{tabular}{l|ccc}
      \centering Update cycle& $AP$ & $AP_{50}$ & $AP_{75}$ \\
     \Xhline{1pt}
 \multirow{1}{*}{Step}         & \multicolumn{1}{c}{27.3}&\multicolumn{1}{c}{45.2}&\multicolumn{1}{c}{28.4}\\
 \hline     
  \multirow{1}{*}{Epoch}         & \multicolumn{1}{c}{\bf27.6}&\multicolumn{1}{c}{\bf45.6}&\multicolumn{1}{c}{\bf28.7}\\
     \end{tabular}
     \end{center}
     \caption{Results on different update cycle for class balanced term.}
     \label{tab:hptab7}
 \end{table}

\subsection{The Effect of Schedules}
{\bf Learning Rate Schedule.}
As shown in Table \ref{tab:hptab6}, Our method has a more benefit when using cosine annealing than step decay as learning rate schedule. With cosine annealing, there is a performance improvement of +1.2 AP. But using step decay as the original setting {\cite{zhou2019objects}} can only improve +0.5 AP. As the epoch increases, the scale of weighting factor approaches inverse class frequency, and it may become so larger or smaller. At this time, the learning rate must also be reduced accordingly so that parameters of the model are not much distracted from the original parameters. For this reason, we guess cosine annealing schedule works well with our method. 

{\bf Update Period of Class Balanced Term.}
Class balanced term varies as training progresses when using linear scheduling. Thus, we report the result on different update cycle for class balanced term changed linearly. In terms of performance, updating the class balanced term per iteration is worse than updating it per epoch as shown in Table \ref{tab:hptab7}. The re-weighting method is a rather unstable method that changes the geometry of the loss surface, so maintaining it consistently for at least an epoch can make training stable. We use cosine annealing as learning rate schedule and update class balanced term for each epoch in all experiments unless noted.

 \begin{table}[!t]
     \begin{center}
     \begin{tabular}{l|ccc}
      \centering & $AP$ & $AP_{50}$ & $AP_{75}$\\
     \Xhline{1pt}
 \multirow{1}{*}{Sigmoid}         & \multicolumn{1}{c}{26.0}&\multicolumn{1}{c}{44.1}&\multicolumn{1}{c}{26.8}\\
  \multirow{1}{*}{+BOFL}         & \multicolumn{1}{c}{\bf26.6}&\multicolumn{1}{c}{\bf44.9}&\multicolumn{1}{c}{\bf27.4}\\
 \hline     
 \multirow{1}{*}{Focal loss}         & \multicolumn{1}{c}{26.4}&\multicolumn{1}{c}{43.4}&\multicolumn{1}{c}{27.4}\\
  \multirow{1}{*}{+BOFL}         & \multicolumn{1}{c}{\bf27.6}&\multicolumn{1}{c}{\bf45.6}&\multicolumn{1}{c}{\bf28.7}\\
     \end{tabular}
     \end{center}
     \caption{The result on sigmoid cross entropy loss function.}
     \label{tab:hptab8}
 \end{table}

\subsection{Sigmoid Cross Entropy Loss Function}
We experiment our method on sigmoid cross entropy. When using sigmoid cross entropy, the scale of the loss at the beginning of training becomes much larger due to many background samples. So we use a smaller value with \(\pi\) = 0.005 for fair comparison when bias of the last convolution in the classification head is initialized as \(b\) = $-$log$((1-$\(\pi\)$) /$ \(\pi\)$)$. As shown in Table \ref{tab:hptab8}, Focal loss {\cite{lin2017focal}} can improve AP more than sigmoid because it reduces the influence of many easy samples in the loss. Therefore, it helps the network
learn with re-weighting efficiently by focusing on samples that they need to learn. This means that our method works best with focal loss and it can be viewed as optimized version for \(\alpha\)-balanced term in focal loss.

 \begin{table}[!t]
     \begin{center}
     \begin{tabular}{l|ccc}
      \centering & $AP$ & $AP_{50}$ & $AP_{75}$\\
     \Xhline{1pt}
 \multirow{1}{*}{FCOS R50}         & \multicolumn{1}{c}{36.4}&\multicolumn{1}{c}{55.5}&\multicolumn{1}{c}{38.7}\\
  \multirow{1}{*}{+BOFL}         & \multicolumn{1}{c}{\bf36.8}&\multicolumn{1}{c}{\bf56.2}&\multicolumn{1}{c}{\bf39.2}\\
 \hline     
 \multirow{1}{*}{FCOS MV2}         & \multicolumn{1}{c}{30.4}&\multicolumn{1}{c}{47.5}&\multicolumn{1}{c}{32.1}\\
  \multirow{1}{*}{+BOFL}         & \multicolumn{1}{c}{\bf31.1}&\multicolumn{1}{c}{\bf48.6}&\multicolumn{1}{c}{\bf32.8}\\
     \end{tabular}
     \end{center}
     \caption{Results on different anchor free detector framework. 'R50' and 'MV2' stand for Resnet 50 and Mobilenet v2 as backbone, respectively.}
     \label{tab:hptab9}
 \end{table}
 
\subsection{Anchor Free Detection Framework}
We report the result on different anchor free detection framework known as FCOS {\cite{tian2019fcos}}. It is anchor free detector using focal loss and utilizes centerness prediction branch to suppress false positives. We train with 16 batches for 12 epochs and set the deferred epoch to 3 with cosine annealing learning rate schedule. We use larger \(\eta\) because FCOS takes mini-batch level balancing (4 images) while CenterNet takes sing-batch level balancing (1 image). It means summing the number of objects in mini-batch level is greater than that of single batch. Through a simple hyper-parameter search, we set \(\eta\) = 0.999. Other settings follow its original implementation {\cite{tian2019fcos}}. As shown in Table \ref{tab:hptab9}, our method works well with FCOS as well.

 \begin{table}[!t]
     \begin{center}
     \begin{tabular}{l|ccc}
      \centering & $AP$ & $AP_{50}$ &$ AP_{75}$\\
     \Xhline{1pt}
 \multirow{1}{*}{Baseline}         & \multicolumn{1}{c}{26.4}&\multicolumn{1}{c}{43.4}&\multicolumn{1}{c}{27.4}\\
  \hline 
  \multirow{1}{*}{CBL}         & \multicolumn{1}{c}{25.1}&\multicolumn{1}{c}{41.3}&\multicolumn{1}{c}{26.1}\\
 \multirow{1}{*}{+linear}         & \multicolumn{1}{c}{26.5}&\multicolumn{1}{c}{44.0}&\multicolumn{1}{c}{27.3}\\
  \multirow{1}{*}{+approx.}         & \multicolumn{1}{c}{26.6}&\multicolumn{1}{c}{44.4}&\multicolumn{1}{c}{27.7}\\
  \multirow{1}{*}{+class-wise}         & \multicolumn{1}{c}{27.2}&\multicolumn{1}{c}{45.1}&\multicolumn{1}{c}{28.1}\\
 \hline  
   \multirow{1}{*}{EQL}         & \multicolumn{1}{c}{25.7}&\multicolumn{1}{c}{42.4}&\multicolumn{1}{c}{26.8}\\
 \hline  
   \multirow{1}{*}{Ours}         & \multicolumn{1}{c}{\bf27.6}&\multicolumn{1}{c}{\bf45.6}&\multicolumn{1}{c}{\bf28.7}\\
     \end{tabular}
     \end{center}
     \caption{Comparison with existing methods. 'linear', 'approx.' and 'class-wise' stand for using linear scheduling technique (Equation \(\ref{eq:hp6}\)),  approximated weighting factor (Equation \(\ref{eq:hp5}\)) and class-wise re-weighting (Equation \(\ref{eq:hp9}\)), respectively.}
     \label{tab:hptab10}
 \end{table}
 
 \begin{table}[!t]
     \begin{center}
     \begin{tabular}{l|c|c}
      \centering & $AP_{tail}$ & $AP_{head}$\\
     \Xhline{1pt}
 \multirow{1}{*}{Baseline}         & \multicolumn{1}{c|}{\bf27.9}&\multicolumn{1}{c}{25.8}\\
  \hline 
   \multirow{1}{*}{EQL}         & \multicolumn{1}{c|}{25.4}&\multicolumn{1}{c}{25.8}\\
     \end{tabular}
     \end{center}
     \caption{Comparison by tail and head classes on MS-COCO. Classes are divided based on Tail Ratio of 8.67\% (image-centric) by following optimal value found in {\cite{tan2020equalization}}.}
     \label{tab:hptab11}
 \end{table}
 
\subsection{Comparison with Existing Methods}

Since CenterNet {\cite{zhou2019objects}} uses penalty-reduced pixel wise logistic regression with focal loss as an objective function, values ranging from 0 to 1 are continuously distributed for every pixel on the feature map in ground-truth. Therefore, we treat the zero part in ground-truth as the background and the rest as the foreground.

{\bf Class Balanced Focal Loss.}
As shown in Table \ref{tab:hptab10}, using class balanced focal loss (CBL) {\cite{cui2019class}} degrades performance of -1.3 AP. When linear scheduling is added to solve performance degradation due to re-weighting, it slightly improves the performance compared to the baseline. And using approximated weighting factor does not significantly affect performance as our expected. In addition, we show that applying re-weighting for each class as Equation \ref{eq:hp9} has better performance than applying it for each sample as Equation \ref{eq:hp1}. The weight factor of the background sample is just set to 1 for the experiment.

 {\bf Equalization Loss.}
 We conduct a performance comparison experiment with Equalization loss (EQL) {\cite{tan2020equalization}}. The value between 8 $\sim$ 9\% is used as Tail Ratio for dividing the class groups based on the optimal value found in {\cite{tan2020equalization}}. As shown in Table \ref{tab:hptab10}, performance decreases by -0.7 AP when using EQL, and we expect that the cause is the data without extreme class imbalance. For this reason, assuming a tail class that includes an extreme few shots is broken, and the action to prevent the suppression effect for tail classes causes even more disadvantage than the gain. In fact, AP of the group selected as the tail class is degraded significantly as shown in Table \ref{tab:hptab11} and it supports our hypothesis.

\subsection{Experiments on Long Tailed Dataset}
We report the result on long tailed imbalance data called COCO-LT from {\cite{wang2020devil}}. This is sampled from MS-COCO {\cite{lin2014microsoft}}. It has extremely imbalanced distribution among different foreground categories such as LVIS {\cite{gupta2019lvis}}. According to Table \ref{tab:hptab12}, the performance for anchor free detector with EQL is -0.8 AP lower than the baseline. EQL {\cite{tan2020equalization}} assumed a second stage classifier {\cite{he2017mask}} that receives balanced samples that keep the ratio between foreground and background class similar. However, the anchor free object detector has a large number of background samples, so training background samples itself can give the suppression effect to the tail classes and it leads to performance degradation. It is supported by the experiment result of Table \ref{tab:hptab12} that performance increases by simply lowering the loss weight of tail classes for the background samples by half.

Our method significantly degrades performance by -3.3 AP on the extremely long tailed dataset. This is because weighting up the loss of the minority classes very much may cause instability in optimization {\cite{cao2019learning}}. We can improve the performance by reducing the intensity of the weighting factor in the dataset level to 5\% as Equation \ref{eq:hp6}. However, since this is beyond the scope of our paper, we leave it as a future work.

 \begin{table}[!t]
     \begin{center}
     \begin{tabular}{l|ccc}
      & $AP$ & $AP_{50}$ & $AP_{75}$ \\
     \Xhline{1pt}
 \multirow{1}{*}{Baseline}         & \multicolumn{1}{c}{14.1}&\multicolumn{1}{c}{25.4}&\multicolumn{1}{c}{13.9}\\
  \hline 
  \multirow{1}{*}{EQL}         & \multicolumn{1}{c}{13.3}&\multicolumn{1}{c}{24.1}&\multicolumn{1}{c}{13.1}\\
 \multirow{1}{*}{EQL (bg. 50\%)}         & \multicolumn{1}{c}{\bf13.6}&\multicolumn{1}{c}{\bf24.6}&\multicolumn{1}{c}{\bf13.3}\\
  \hline  
  \multirow{1}{*}{Ours}         & \multicolumn{1}{c}{10.8}&\multicolumn{1}{c}{20.7}&\multicolumn{1}{c}{10.0}\\
   \multirow{1}{*}{Ours (intensity 5\%)}         & \multicolumn{1}{c}{\bf15.0}&\multicolumn{1}{c}{\bf27.0}&\multicolumn{1}{c}{\bf14.9}\\
     \end{tabular}
     \end{center}
     \caption{Results on long tailed dataset COCO-LT. 'bg. 50\%' indicates lowering the loss weight of tail classes for the background samples to 50\%. 'intensity 5\%' stands for reducing the intensity of the weighting factor in the dataset level to 5\%.}
     \label{tab:hptab12}
 \end{table}
 
\section{Conclusion}
In this paper, we have presented Balance-oriented focal loss to train a general unbalanced data for anchor free detectors. In order to solve the background class imbalance problem as well as foreground class imbalance problem simultaneously, we optimizes \(\alpha\)-balanced term of the focal loss for each batch and for each epoch. It can improve +1.2 AP in MS-COCO for the anchor free real-time detector. Our method is a simple and easy to apply. And we have analyzed effectiveness of our components through extensive experiments.

{\small
\bibliographystyle{ieee_fullname}
\bibliography{egbib}
}

\end{document}